\documentclass[twoside]{article}

\usepackage[sc]{mathpazo} 
\usepackage[T1]{fontenc} 
\usepackage{microtype} 

\usepackage[hmarginratio=1:1,top=32mm,columnsep=20pt]{geometry} 
\usepackage{multicol} 
\usepackage[hang, small,labelfont=bf,up,textfont=it,up]{caption} 
\usepackage{booktabs} 
\usepackage{float} 
\usepackage{hyperref} 

\usepackage{lettrine} 
\usepackage{paralist} 

\usepackage{abstract} 

\usepackage{titlesec} 
\renewcommand\thesection{\Roman{section}}
\titleformat{\section}[block]{\large\scshape\centering}{\thesection.}{1em}{} 

\usepackage{fancyhdr} 

\usepackage{amssymb}
\usepackage{epsfig}
\usepackage{amssymb}
\usepackage{url}
\usepackage{amsmath}
\usepackage{natbib}
\usepackage{amsthm}
\usepackage{booktabs}
\usepackage[lined,boxed,linesnumbered]{algorithm2e}
\usepackage{url}
\usepackage{alltt}
\usepackage{enumerate}
\usepackage{multirow}
\usepackage{setspace}
\usepackage{bm}
\usepackage{color}

\newcommand{\nmathbf}{\bm}

\def\bfA{\nmathbf A}

\def\bfI{\nmathbf I}

\def\bfW{\nmathbf W}
\def\bfX{\nmathbf X}
\def\bfY{\nmathbf Y}
\def\bfZ{\nmathbf Z}

\def\bfu{\nmathbf u}

\def\bfbeta   {\nmathbf \beta}

\def\bfeta    {\nmathbf \eta}
\def\bftheta  {\nmathbf \theta}

\def\bfmu     {\nmathbf \mu}

\def\bfSigma  {\nmathbf \Sigma}

\newcommand{\bfone}{{\nmathbf 1}}

\def\boldfacefake#1{\kern-4pt
    \hbox{ \mathsurround=0pt
    \hbox to 0.4pt{$#1$\hss}\hbox to 0.4pt{$#1$\hss}\hbox {$#1$}}}

\newcommand{\N}{\mbox{N}}

\newcommand{\be}{\begin{eqnarray}}
\newcommand{\ee}{\end{eqnarray}}
\newcommand{\ba}{\begin{eqnarray*}}
\newcommand{\ea}{\end{eqnarray*}}

\newtheorem{theorem0}{Theorem}
\newtheorem{lemma0}{Lemma}
\newtheorem{remark0}{Remark}
\newtheorem{fact0}{Fact}
\newtheorem{example0}{Example}
\newtheorem{definition0}{Definition}
\newtheorem{corollary0}{Corollary}
\newtheorem{proposition0}{Proposition}
\newtheorem{algorithmY}{Algorithm}
\newtheorem{conjecture0}{Conjecture}

\title{\vspace{-15mm}\fontsize{19pt}{10pt}\selectfont\textbf{Probit Normal Correlated Topic Models}} 

\author{
\textbf{Xingchen Yu}\\
\normalsize Rochester Institute of Technology \\ 
\normalsize 98 Lomb Memorial Drive, Rochester, NY 14623, USA \\ 
\normalsize \href{mailto:xvy5021@gmail.com}{xvy5021@gmail.com} 
\vspace{5mm}
\and
\textbf{Ernest Fokou\'e}\\
\normalsize Rochester Institute of Technology \\ 
\normalsize 98 Lomb Memorial Drive, Rochester, NY 14623, USA \\ 
\normalsize \href{mailto:ernest.fokoue@rit.edu}{ernest.fokoue@rit.edu} 
\vspace{-5mm}
}
\date{}

\begin{document}

\maketitle 

\thispagestyle{fancy} 

\begin{abstract}
The logistic normal distribution has recently been adapted via the transformation of multivariate Gaussian variables to model the topical distribution of documents in the presence of correlations among topics. In this paper, we propose a probit normal alternative approach to modelling correlated topical structures. Our use of the probit model in the context of topic discovery is novel, as many authors have so far concentrated solely of the logistic model partly due to the formidable inefficiency of the multinomial probit model even in the case of very small topical spaces. We herein circumvent the inefficiency of multinomial probit estimation by using an adaptation of the diagonal orthant multinomial probit in the topic models context, resulting in the ability of our topic modelling scheme to handle corpuses with a large number of latent topics. An additional and very important benefit of our method
lies in the fact that unlike with the logistic normal model whose non-conjugacy leads to the need for sophisticated sampling schemes, our approach exploits the natural conjugacy inherent in the auxiliary formulation of the probit model to achieve greater simplicity.
The application of our proposed scheme to a well known Associated Press corpus not only helps discover a large number of meaningful topics but also reveals the capturing of compellingly intuitive correlations among certain topics. Besides, our proposed approach lends itself to even further scalability thanks to various existing high performance algorithms and architectures capable of handling millions of documents.
\end{abstract}




\section{Introduction}
The task of recovering the latent topics underlying a given corpus of $D$ documents has been in the forefront of active research in statistical
machine learning for more than a decade, and continues to receive the dedicated contributions from many researchers from around the world.
Since the introduction of Latent Dirichlet Allocation (LDA) \cite{Blei03latentdirichlet} and then the extension to correlated topic models (CTM) \cite{Blei06correlatedtopic}, a series of excellent contributions have been made to this exciting field, ranging from slight extension
in the modelling structure to the development of scalable topic modeling algorithms capable of handling extremely large
collections of documents, as well as selecting an optimal model among a collection of competing models or using the output of topic modelling as entry points (inputs)
to other machine learning or data mining tasks such as image analysis and sentiment extraction, just to name a few. As far as correlated topic models are concerned,
virtually all the contributors to the field have so far concentrated solely on the use of the logistic normal topic model. The seminal paper on correlated topic model\cite{Blei06correlatedtopic} adopts a variational approximation approach to model fitting while subsequent authors like \cite{Mimno_gibbssampling} propose a Gibbs sampling scheme with data augmentation of uniform random variables. More recently, \cite{Tsinghua:2013:1} presented an exact and scalable Gibbs sampling algorithm with Polya-Gamma distributed auxiliary variables which is a recent development of efficient sampling of logistic model. Despite the inseparable relationship between logistic and probit model in statistical modelling, the probit model has not yet been proposed, probably due to its computational inefficiency for multiclass classification problem and high posterior dependence between auxiliary variables and parameters. As for practical application where topic models are commonly employed, having multiple topics is extremely prevalent. In some cases, more than 1000 topics will be fitted to large datasets such as Wikipedia and Pubmed data. Therefore, using MCMC probit model in topic modeling application will be impractical and inconceivable due to its computational inefficiency. Nonetheless, a recent work on diagonal orthant probit model \cite{Johndrow:2013:1} substantially improved the sampling efficiency while maintaining the predictive performance, which motivated us to build an alternative correlated topic modeling with probit normal topic distribution. On the other hand,
probit models inherently capture a better dependency structure between topics and co-occurrence of words within a topic as it doesn't assume the IIA (independence of irrelevant alternatives) restriction of logistic models.\\
\\
The rest of this paper is organized as follows: in section 2, we present a conventional formulation of topic modelling along with our general notation and the correlated topic models
extension.
Section 3 introduces our adaptation of the diagonal orthant probit model to topic discovery in the presence correlations among topics, along with the corresponding auxiliary variable sampling scheme for updating the probit model parameters and the remainder of all the posterior distributions of the parameters of the model. Unlike with the logistic normal formulation
where the non-conjugacy leads to the need for sophisticated sampling scheme, in this section we clearly reveal the simplicity of our proposed method resulting from the natural conjugacy inherent in the auxiliary formulation of the updating of the parameters. We also show compelling computational demonstrations of the efficiency of the diagonal orthant approach compared to the traditional multinomial probit for on both the auxiliary variable sampling and the estimation of the topic distribution.
Section 4 presents the performance of our proposed approach on the Associated Press data set, featuring the intuitively appealing topics discovered, along with
the correlation structure among topics and the loglikelihood as a function of topical space dimension. Section 5 deals with our conclusion, discussion
and elements of our future work.

\section{General aspects of topic models}
In a given corpus, one could imagine that each document deals with one or more topics. For instance, one of the collection considered
in this paper is provided by the Associated Press and covers topics as varied as {\it aviation, education, weather, broadcasting, air force, navy,
national security, international treaties, investing, international trade, war, courts, entertainment industry, politics}, and etc.
From a statistical perspective, a topic is often modeled as a {\it probability distribution over words}, and as a result a given document
is treated as a {\it mixture of probabilistic topics} \cite{Blei03latentdirichlet}. We consider a setting where we have a total of $V$ unique words in the reference vocabulary and $K$ topics underlying the $D$ documents provided. Let $w_{dn}$ denote the $n$-th word in the $d$-th document, and let $z_{dn}$ refer to the label of the topic assigned to the $n$-th word of that $d$-th document. Then the probability of $w_{dn}$ is given by

\begin{eqnarray}
\Pr(w_{dn}) = \sum_{k=1}^K{\Pr(w_{dn}| z_{dn}=k)\Pr(z_{dn}=k)},
\label{eq:tm:1}
\end{eqnarray}
where $\Pr(z_{dn}=k)$ is the probability that the $n$th word in the $d$th document is assigned to topic $k$.
This quantity plays an important role in the analysis of correlated topic models. In the seminal article on correlated
topic models \cite{Blei06correlatedtopic}, $\Pr(z_{dn}=k)$ is modeled for each document $d$ as a function of a $K$-dimensional vector
$\bfeta_d$ of parameters. Specifically, the logistic-normal defines $\bfeta_d = (\eta_{d}^1, \eta_{d}^2, \cdots, \eta_{d}^K)$ where the last element $\eta_{d}^K$ is typically set to zero for identifiability and assumes
with  $\bfeta_d \sim {\tt MVN}(\bfmu, \bfSigma)$ with
$$
\theta_{d}^k =  \Pr[z_{dn}^k = 1 | \bfeta_d] =f(\bfeta_d) = \frac{e^{\eta_{d}^k}}{\sum_{j=1}^K{e^{\eta_{d}^j}}},\quad k=1,2,\cdots,K-1 \quad \textit{and} \quad \theta_{d}^K = \frac{1}{\sum_{j=1}^K{e^{\eta_{d}^j}}},
$$
Also, $\forall n \in \{1,2, \cdots, N_d\}\,\,\,$ and $z_{dn} \sim {\tt Mult}(\bftheta_{d})$, and
$w_{dn} \sim {\tt Mult}(\bfbeta)$. With all these model components defined, the estimation task in correlated topic modelling from a Bayesian perspective
can be summarized in the following posterior
\begin{eqnarray}
p(\bfeta_d,\bfZ |\bfW, \bfmu, \bfSigma) &\propto& p(\bfW|\bfZ)\prod_{d=1}^D\left\{\prod_{n=1}^{N_d} p(z_{dn}) p(\bfeta_d|\bfmu, \bfSigma)\right\}\\ \nonumber
&=&\prod_{k=1}^{K}{\frac{\delta(C_{k}+ \bfbeta)}{\delta(\bfbeta)}}\prod_{d=1}^D\left\{\left(\prod_{n=1}^{N_d} \theta_d^{z_{dn}}\right) \mathcal{N}(\bfeta_d|\bfmu, \bfSigma)\right\},
\label{eq:post:1}
\end{eqnarray}
where $\delta(\cdot)$ is defined using the Gamma function $Gamma(\cdot)$ so for a $K$-dimension vector $\bfu$,
$$
\delta(\bfu) = \displaystyle \frac{\displaystyle \prod_{k=1}^K{\Gamma(u_k)}} {\Gamma\left(\displaystyle \sum_{k=1}^K{u_k}\right)}.
$$
\eqref{eq:post:1} provides the ingredients for estimating the parameter vectors $\bfeta_d$ that help capture the correlations among topics, and the matrix $\bfZ$
that contains the topical assignments. Under the logistic normal model, sampling from the full posterior of $\bfeta_d$ derived from the joint posterior in \eqref{eq:post:1} requires the use of sophisticated sampling schemes like the one used in \cite{Tsinghua:2013:1}. Although these authors managed to achieve great performances on large corpuses of documents, we thought it useful to contribute to correlated topic modelling by way of the multinomial probit. Clearly, as indicated earlier, most authors concentrated on logistic-normal even despite non-conjugacy, and the lack of probit topic modeling can be easily attributed to the inefficiency of the corresponding sampling scheme. In the most raw formulation of the multinomial probit that intends to capture the full extend of all the correlations among the topics, the topic assignment probability is defined by \eqref{eq:probit:0}.
\begin{eqnarray}
\Pr(z_{dn}=k)=\theta_{d}^k =  \int\int\int\cdots\int{\phi_{K}(u; \bfeta_d, R) d u}
\label{eq:probit:0}
\end{eqnarray}
The practical evaluation of \eqref{eq:probit:0} involves a complicated high dimensional integral which is typically computationally intractable when the number of categories is greater than $4$. A relaxed version of \eqref{eq:probit:0}, one that still captures more correlation than the logit and that is also very commonly used in practice, defines $\theta_d^k$ as
\begin{eqnarray}
\theta_{d}^k = \int_{-\infty}^{+\infty}{\left\{\prod_{j=1, j\neq k}^{K}{\Phi(v + \eta_{d}^k - \eta_{d}^{j})}\right\}\phi(v) dv}
= \mathbb{E}_{\phi(v)}\left\{\prod_{j=1, j\neq k}^{K}{\Phi(V + \eta_{d}^k - \eta_{d}^{j})}\right\},
\label{eq:probit:prob:1}
\end{eqnarray}
where $\phi(v)=\frac{1}{\sqrt{2\pi}}e^{-\frac{1}{2}v^2}$ is the standard normal density, and $\Phi(v)=\int_{-\infty}^{v}{\phi(u)du}$ is the standard normal distribution function.
Despite this relaxation, the multinomial probit in this formulation still has major drawbacks namely: (a) Even when one is given the vector $\bfeta_d$, the calculation of
$\theta_d^k$ remains computationally prohibitive even for moderate values of $K$. In practice, one may consider using a monte carlo approximation to that integral in \eqref{eq:probit:prob:1}. However, such an approach in the context of a large corpus with many underlying latent topics renders the probit formulation almost unusable.
(b) As far as the estimation of $\bfeta_d$ is concerned, a natural approach to sampling from the posterior of $\bfeta_d$ in this context would be to use the Metropolis-Hastings
updating scheme, since the full posterior in this case is not available. Unfortunately, the Metropolis in this case is excruciatingly slow with poor mixing rates and high sensitivity to the proposal distribution. It turns out that an apparently appealing solution in this case could come from the auxiliary variable formulation as described in \cite{AlbertChib93}. Unfortunately, even this promising formulation fails catastrophically for moderate values $K$ as we will demonstrate in the subsequent section, due to the high dependency structure between auxiliary variables and parameters. Essentially, the need for Metropolis is avoided by defining an auxiliary vector $Y_d$ of dimension $K$. For $n=1,\cdots,N_d$, we consider the vector $z_{dn}$ containing the current topic allocation and we repeatedly sample $Y_{dn}$ from a $K$-dimensional multivariate Gaussian
until the component of $Y_{dn}$ that corresponds to the non-zero index in $z_{dn}$ is the largest of all the components of $Y_{dn}$, ie.
\begin{eqnarray}
Y_{dn}^{z_{dn}} = \underset{k=1,\cdots,K}{\max}\{Y_{dn}^k\}.
\label{eq:max:aux:1}
\end{eqnarray}
The condition in \eqref{eq:max:aux:1} typically fails to be fulfilled even when $K$ is moderately large. In fact, we demonstrate later that in some cases, it becomes impossible to
find a vector $Y_{dn}$ satisfying that condition. Besides, the dependency of $Y_{dn}$ on the current value of $\bfeta_{d}$ further complicates the sampling scheme especially in the case of large topical space. In the next section,
we remedy these inefficiencies by proposing and developing our adaptation of the diagonal orthant multinomial probit.

\section{Diagonal Orthant Probit for Correlated Topic Models}
In a recent work, \cite{Johndrow:2013:1} developed the diagonal orthant probit approach to multicategorical classification. Their approach circumvents the bottlenecks mentioned earlier and substantially improves the sampling efficiency while maintaining the predictive performance. Essentially, the diagonal orthant probit approach successfully makes the most of the benefits of binary classification, thereby substantially reducing the high dependency that made the condition \eqref{eq:max:aux:1} computationally unattainable. Indeed, with the diagonal orthant multinomial model, we achieved three main benefits
\begin{itemize}
\item A  more tractable and easily computatble definition of topic distribution $\theta_d^k=\Pr(z_{dn}=k|\bfeta_d)$
\item A clear and very straightforward and adaptable auxiliary variable sampling scheme
\item The capacity to handle a very large number of topics due to the efficiency and low dependency.
\end{itemize}
Under the diagonal orthant probit model, we have
\begin{eqnarray}
\theta_{d}^k = \frac{(1-\Phi(-\eta_d^k))\prod_{j\neq k}\Phi(-\eta_d^{j})}{\displaystyle \sum_{\ell=1}^{K}{(1-\Phi(-\eta_d^\ell))\prod_{j\neq \ell}\Phi(-\eta_d^{j})}}.
\label{eq:orthant:prob:1}
\end{eqnarray}

The generative process of our probit normal topic models is essentially identical to logistic topic models except that the topic distribution for each document now is obtained by a probit transformation of a multivariate Gaussian variable \eqref{eq:orthant:prob:1}. As such, the generating process of a document of length $\textit\N_d $ is as follows:
\begin{enumerate}
  \item Draw $\eta \sim {\tt MVN}(\mu, \Sigma)$ and transform $\eta_{d}$ into topic distribution $\theta_{d}$ where each element of $\theta$ is computed as follows:
  \begin{eqnarray}
\theta_{d}^k = \frac{(1-\Phi(-\eta_d^k))\prod_{j\neq k}\Phi(-\eta_d^{j})}{\displaystyle \sum_{\ell=1}^{K}{(1-\Phi(-\eta_d))\prod_{j\neq \ell}\Phi(-\eta_d^{j})}}.
\end{eqnarray}
  \item For each word position $\textit{n}$ $\in (1,\cdots, \textit\N_d )$
  \begin{enumerate}
    \item Draw a topic assignment $Z_n \sim {\tt Mult} (\theta_d)$
    \item Draw a word $W_n \sim {\tt Mult} (\varphi^{z_n})$
  \end{enumerate}
\end{enumerate}
Where $\Phi(\cdot)$ represents the cumulative distribution of the standard normal.
We specify a Gaussian prior for $\bfeta_d$, namely $(\bfeta_d | \cdots) \sim {\cal N}_{K}(\bfmu, \bfSigma)$.
Throughout this paper, we'll use $\phi_{K}(\cdot)$ to denote the $K$-dimensional multivariate Gaussian density function,
$$
\phi_{K}(\bfeta_d; \bfmu, \bfSigma) = \frac{1}{\sqrt{(2\pi)^K |\Sigma|}}\exp\left\{-\frac{1}{2}(\bfeta_d-\bfmu)^\top\bfSigma^{-1} (\bfeta_d-\bfmu)\right\}.
$$
To complete the Bayesian analysis of our probit normal topic model, we need to sample from the joint posterior
\begin{eqnarray}
p(\bfeta_d, \bfZ_d | \bfW, \bfmu, \bfSigma)  \propto p(\bfeta_d | \bfmu, \bfSigma)p(\bfZ_d | \bfeta_d)p(\bfW|\bfZ_d).
\label{eq:post:joint:2}
\end{eqnarray}
As noted earlier, the second benefit of the diagonal orthant probit model lies in its clear, simple, straightforward yet powerful auxiliary variable sampling scheme.
We take advantage of that diagonal orthant property when dealing with the full posterior for $\bfeta_d$ given by
\begin{eqnarray}
p(\bfeta_d | \bfW, \bfZ_d, \bfmu, \bfSigma)  \propto p(\bfeta_d | \bfmu, \bfSigma)p(\bfZ_d | \bfeta_d).
\label{eq:post:full:eta:1}
\end{eqnarray}
While sampling directly from \eqref{eq:post:full:eta:1} is impractical, defining a collection of auxiliary variables $\bfY_d$
allows a scheme that samples from the joint posterior $p(\bfeta_d,  \bfZ_d, \bfY_d|\bfW,  \bfmu, \bfSigma)$  using the following:
For each document $d$, the matrix $\bfY_d \in \mathbb{R}^{N_d \times K}$ contains all the values of the auxiliary variables,
$$
\bfY_d = \left[\begin{array}{cccccc}
Y_{d1}^1 & Y_{d1}^2 & \cdots & Y_{d1}^k & \cdots & Y_{d1}^K\\
 Y_{d2}^1 & Y_{d2}^2 & \cdots & Y_{d2}^k & \cdots & Y_{d2}^K\\
\vdots & \vdots & \cdots & \ddots & \cdots & \vdots\\
Y_{d,{N_d-1}}^1 & Y_{d,{N_d-1}}^2 & \cdots & Y_{d,{N_d-1}}^k & \cdots & Y_{d,{N_d-1}}^K\\
Y_{d,{N_d}}^1 & Y_{d,{N_d}}^2 & \cdots & Y_{d,{N_d}}^k & \cdots & Y_{d,{N_d}}^K
\end{array}\right]
$$
Each row $Y_{dn} = (Y_{dn}^1, \cdots, Y_{dn}^k, \cdots,  Y_{dn}^K)^\top$ of $\bfY_d$ has $K$ components, and the diagonal orthant
updates them readily using the following straightforward sampling scheme: Let $k$ be the current topic allocation for the nth word.
\begin{itemize}
\item For the component of $Y_{dn}$ whose index corresponds to the label of current topic assignment of word $n$
sample from a truncated normal distribution with variance $1$ restricted to positive outcomes
$$
(Y_{dn}^{k} | \eta_d^{k}) \sim  \mathcal{N}_{+}(\eta_d^{k}, 1) \quad z_{dn}^k = 1
$$
 \item For all components of $Y_{dn}$ whose indices do correspond to the label of current topic assignment of word $n$
sample from a truncated normal distribution with variance $1$ restricted to negative outcomes
$$
(Y_{dn}^{j} | \eta_d^{j}) \sim  \mathcal{N}_{-}(\eta_d^{j}, 1) \quad \quad z_{dn}^{j}\neq 1
$$
\end{itemize}
Once the matrix $\bfY_d$ is obtained, the sampling scheme updates the parameter vector $\bfeta_d$ by conveniently drawing
$$
(\bfeta_d | \bfY_d, \bfA, \bfmu, \bfSigma) \sim MVN(\bfmu_{\bfeta_d}, \bfSigma_{\bfeta_d}),
$$
where
$$
\bfmu_{\bfeta_d} = \bfSigma_{\bfeta_d}(\bfSigma^{-1} \bfmu + \bfX_d^\top \bfA^{-1}{\rm vec}(\bfY_d))
\quad
\text{and}
\quad
\bfSigma_{\bfeta_d} = (\bfSigma^{-1} + \bfX_d^\top \bfA^{-1} \bfX_d)^{-1}.
$$
with $\bfX_d = \bfone_{N_d} \otimes \bfI_K$ and ${\rm vec}(\bfY_d)$ representing the row-wise vectorization of
the matrix $\bfY_d$.
Adopting the fully Bayesian treatment of our probit normal correlated topic model, we add an extra layer to the hierarchy in order to capture
the variation in the mean vector and the variance-covariance matrix of the parameter vector $\bfeta_d$. Taking advantage of
conjugacy, we specify a normal-Inverse-Wishart prior for $(\bfmu, \bfSigma)$, namely,
$$
p(\bfmu, \bfSigma) = NIW(\bfmu_0, \kappa_0, \Psi_0, \nu_0),
$$
meaning that $\bfSigma|\nu_0, \Psi_0 \sim IW(\Psi_0, \nu_0)$ and $(\bfmu | \bfmu_0, \bfSigma, \kappa_0) \sim MVN(\bfmu_0, \Sigma/\kappa_0)$.
The corresponding posterior is normal-inverse-Wishart, so that we can write
$$
p(\bfmu, \bfSigma| \bfW, \bfZ, \bfeta) = NIW(\bfmu^\prime, \kappa^\prime, \Psi^\prime, \nu^\prime),
$$
where $\kappa^\prime=\kappa_0+D$, $\nu^\prime = \nu_0 + D$, $\bfmu^\prime = \frac{D}{D+\kappa_0}\bar{\bfeta} + \frac{\kappa_0}{D+\kappa_0}\bfmu_0$,
and
$$
\Psi^\prime = \Psi_0 + Q + \frac{\kappa_0}{\kappa_0+D}(\bar{\bfeta}-\bfmu_0)(\bar{\bfeta}-\bfmu_0)^\top,
$$
where
$$
Q = \sum_{d=1}^{D}{(\bfeta_d-\bar{\bfeta})(\bfeta_d-\bar{\bfeta})^\top}.
$$

As far as sampling from the full posterior distribution of $Z_{dn}$ is concerned, we use the expression
$$
\Pr[z_{dn}^k = 1 | \bfZ_{\neg n}, w_{dn}, \bfW_{\neg dn}] \propto p(w_{dn}|z_{dn}^k = 1,\bfW_{\neg dn},\bfZ_{\neg n}) \theta_{d}^k \propto \frac{C_{k,\neg n}^{w_{dn}}+\beta_{w_{dn}}}{\sum_{j=1}^V{C_{k,\neg n}^{j}} + \sum_{j=1}^V{\beta_j}} \theta_{d}^k.
$$
where the use of $C_{\cdot,\neg n}$ is used to indicate that the $n$th is not included in the topic or document under consideration.

\section{Computational results on the Associated Press data}
In this section, we used a famous Associated Press data set from \cite{Grün:Hornik:2011:JSSOBK:v40i13} in R to uncover the word topic distribution, the correlation structure between various topics as well as selecting optimal models. The Associated Press corpus consists of 2244 documents and 10473 words. After preprocessing the corpus by picking frequent and common terms, we reduced the size of the words from 10473 to 2643 for efficient sampling. \\

In our first experimentation, we built a correlated topic modelling structure based on the traditional multinomial
probit and then tested the computational speed for key sampling tasks. The high posterior dependency structure between auxiliary variables and parameters make
multinormal probit essentially unscalable for situations where it is impossible for the sampler to yield a random variate of the auxiliary variable corresponding the current topic allocation label that is also the maximum \eqref{eq:max:aux:1}. For a random initialization of topic assignment, the sampling of auxiliary variable cannot even complete one single iteration. In the case of good initialization of topical prior $\bfeta_d$ which leads to smooth sampling of auxiliary variables, the computational efficiency is still undesirable and we observed that for larger topical space such as K=40, the auxiliary variable stumbled again after some amount of iterations, indicating even good initialization will not ease the troublesome dependency relationship between the auxiliary variables and parameters in larger topical space. Unlike with the traditional probit model for which the computation of $\theta_d^k$ is virtually impractical for large $K$,the diagonal orthant approach makes this computation substantially faster ever for large $K$. The comparison of the computational speed of two essential sampling tasks between the multinomial probit model and digonal orthant probit model are shown as below in table 1 \eqref{tab:comp:efficiency:1}. \\

\begin{table}[!htbp]
\centering
\begin{tabular}{lrr}
\hline
  Sampling Task (K=10)	 & {MNP} &	 {DO Probit} \\
Topic Distribution $\theta$ & $18.3$	& $0.06$ \\
Auxiliary variable $Y_d$ & ($108$ to NA)	& $3.09$ \\
\hline
\end{tabular}

\begin{tabular}{lrr}
\hline
  Sampling Task (K=20)	 & {MNP} &	 {DO Probit} \\
Topic Distribution $\theta$ & $63$	& $0.13$ \\
Auxiliary variable $Y_d$ & ($334$ to NA)	& $3.39$ \\
\hline
\end{tabular}

\begin{tabular}{lrr}
\hline
  Sampling Task (K=30)	 & {MNP} &	 {DO Probit} \\
Topic Distribution $\theta$ & $123$	& $0.21$ \\
Auxiliary variable $Y_d$ & ($528$ to NA)	& $3.49$ \\
\hline
\end{tabular}

\begin{tabular}{lrr}
\hline
  Sampling Task (K=40)	 & {MNP} &	 {DO Probit} \\
Topic Distribution $\theta$ & $211.49$	& $0.33$ \\
Auxiliary variable $Y_d$ & ($1785$ to NA)	& $3.79$ \\
\hline
\end{tabular}
\caption{All the numbers in this table represent the processing time (in seconds), and are computed in R on PC using a parallel algorithm acting on 4 CPU cores.
NA here represents situations where it is impossible for the sampler to yield a random variate of the auxiliary variable corresponding the current topic allocation label that is also the maximum}
\label{tab:comp:efficiency:1}
\end{table}

In addition to the drastic improvement of the overall sampling efficiency, we noticed that the computational complexity for sampling the auxiliary variable and topic distribution is close to O(1) and O(K) respectively, suggesting that probit normal topic model now becomes an attainable and feasible tool of the traditional correlated topic model.\\

\begin{figure}[!htbp]
  \centering
  \epsfig{figure=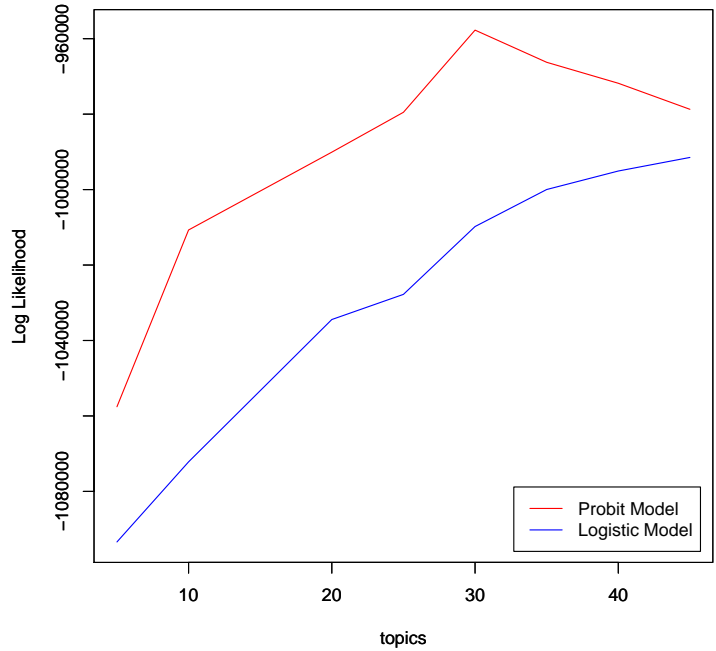, width=10cm, height=10cm}
  \caption{Loglikelihood as a function of the number of topics}
  \label{fig:loglikelihood:full:1}
\end{figure}

Central to topic modelling is the need to determine for a given corpus the optimal number of latent
topics. As it is the case for most latent variable models, this task can be formidable at times, and there
is no consensus among machine learning researchers as to which of the existing methods is the best.
Figure \eqref{fig:loglikelihood:full:1} shows the loglikelihood as a function of the number of topics
discovered in the model. Apart from the loglikelihood, many other techniques are commonly used such as perplexity,
harmonic mean method and so on.\\

As we see, the optimal number of topics in this case is 30.
In table \eqref{tab:topics:1c}, we show a subset of the 30 topics uncovered where each topic is represented by the 10 most frequent words. It can be seen that our probit normal topic model is able to capture the co-occurrence of words within topics successfully. In figure 2, we also show the correlation structure between various topics which is the essential purpose of employing the correlated topic model. Evidently, the correlation captured intuitively reflect the natural relationship between similar topics.\\

\begin{table}[!htbp]
\centering
\begin{tabular}{llllllll}
  \toprule
 &  {\bf Topic 25} &  {\bf Topic 18} &  {\bf  Topic 23} &  {\bf  Topic 11} &  {\bf Topic 1} &  {\bf Topic 24} &  {\bf Topic 27} \\
  \hline
  {\bf Word1} & court & company & bush & students & tax & fire & air \\
  {\bf Word2} & trial & billion & senate & school & budget & water & plane \\
  {\bf   Word3} & judge & inc & vote & meese & billion & rain & flight \\
  {\bf   Word4} & prison & corp & dukakis & student & bill & northern & airlines \\
  {\bf   Word5} & convicted & percent & percent & schools & percent & southern & pilots \\
  {\bf   Word6} & jury & stock & bill & teachers & senate & inches & aircraft \\
  {\bf   Word7} & drug & workers & kennedy & board & income & fair & planes \\
  {\bf   Word8} & guilty & contract & sales & education & legislation & degrees & airline \\
  {\bf   Word9} & fbi & companies & bentsen & teacher & taxes & snow & eastern \\
  {\bf   Word10} & sentence & offer & ticket & tax & bush & temperatures & airport \\
   \hline
\end{tabular}
\end{table}
\begin{table}[!htbp]
\centering
\begin{tabular}{llllllll}
  \toprule
 & {\bf Topic 6} & {\bf Topic 12} & {\bf Topic 20} & {\bf Topic 2} & {\bf Topic 22} & {\bf Topic 16} & {\bf Topic 15} \\
  \hline
{\bf Word1} & percent & space & military & soviet & aid & police & dollar \\
   {\bf Word2} & stock & shuttle & china & gorbachev & rebels & arrested & yen \\
  {\bf Word3} & index & soviet & chinese & bush & contras & shot & rates \\
  {\bf Word4} & billion & nasa & soldiers & reagan & nicaragua & shooting & bid \\
  {\bf Word5} & prices & launch & troops & moscow & contra & injured & prices \\
  {\bf Word6} & rose & mission & saudi & summit & sandinista & car & price \\
 {\bf Word7} & stocks & earth & trade & soviets & military & officers & london \\
  {\bf Word8} & average & north & rebels & treaty & ortega & bus & gold \\
 {\bf Word9} & points & korean & hong & europe & sandinistas & killing & percent \\
 {\bf Word10} & shares & south & army & germany & rebel & arrest & trading \\
   \hline
\end{tabular}
\end{table}

\begin{table}[!htbp]
\centering
\begin{tabular}{llllllll}
  \toprule
 & {\bf Topic19} & {\bf Topic 14} & {\bf Topic 7} & {\bf Topic 4} & {\bf Topic 30} & {\bf Topic 8} & {\bf Topic 17} \\
  \hline
 {\bf Word1} & iraq & trade & israel & navy & percent & south & film \\
   {\bf Word2} & kuwait & percent & israeli & ship & oil & africa & movie \\
   {\bf Word3} & iraqi & farmers & jewish & coast & prices & african & music \\
   {\bf Word4} & german & farm & palestinian & island & price & black & theater \\
   {\bf Word5} & gulf & billion & arab & boat & cents & church & actor \\
   {\bf Word6} & germany & japan & palestinians & ships & gasoline & pope & actress \\
   {\bf Word7} & saudi & agriculture & army & earthquake & average & mandela & award \\
   {\bf Word8} & iran & japanese & occupied & sea & offers & blacks & band \\
   {\bf Word9} & bush & tons & students & scale & gold & apartheid & book \\
  {\bf Word10} & military & drought & gaza & guard & crude & catholic & films \\
   \hline
\end{tabular}
\caption{Representation of topics discovered by our method}
\label{tab:topics:1c}
\end{table}


\begin{figure}[!htbp]
  \centering
  \epsfig{figure=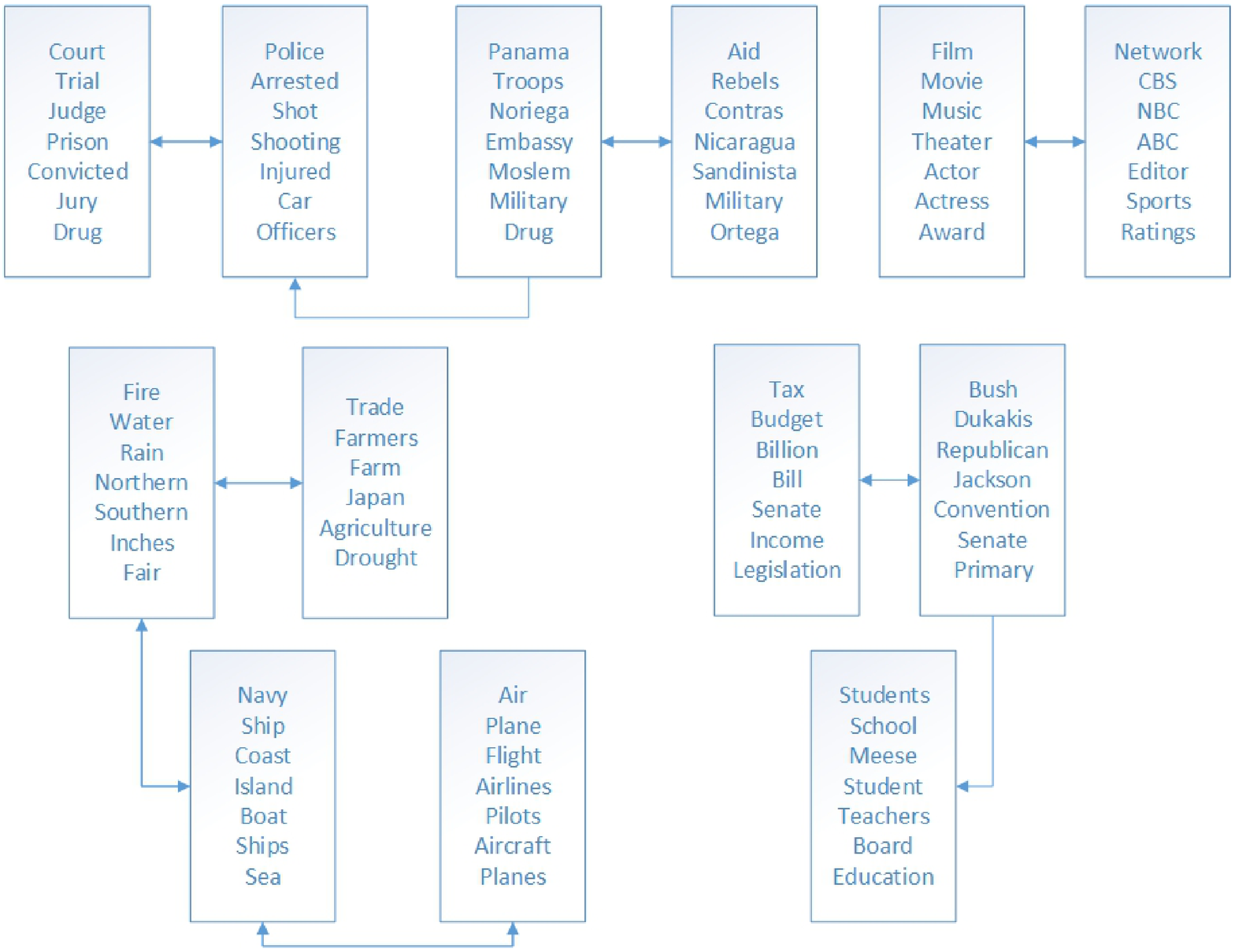, width=10cm, height=8cm}
  \caption{Graphical representation of the correlation among topics}
  \label{fig:correlation:topics:1}
\end{figure}
\newpage
\section{Conclusion and Discussion}
In the context of topic modelling where many other researchers seem to have avoided it. By adapting the diagonal orthant probit model, we proposed a probit alternative to the logit approach to the topic modeling. Compared to the multinomial probit model we constructed, our topic discovery scheme using diagonal orthant probit model enjoyed several desirable properties; First,we gained the efficiency in computing the topic distribution $\bftheta_d^{k}$ ; Second, we achieved a clear and very straightforward and adaptable auxiliary variable sampling scheme that substantially reduced the strength of the dependence structure between auxiliary variables and model parameters, responsible for absorbing state in the Markov chain; Thirdly, as a consequence of good mixing, our approach made the probit model a viable and competitive alternatives to its logistic counterpart. In addition to all these benefits, our proposed method offers a straightforward and inherent conjugacy, which helps avoid those complicated sampling schemes employed in the logistics normal probit model.\\

In the Associated Press example explored in the previous section, not only does our method produce a better likelihood than the logistic normal topic model with variational EM, but also discovers meaningful topics along with underlying correlation structure between topics. Overall, the method we developed in this paper offers another feasible alternatives in the context of correlated topic model that we hope will be further explored and extended by many other researchers\\

Based on the promising results we have seen in this paper, the probit normal topic model opens the door for various future works. For instance, \cite{DBLP:conf:sdm:SalomatinYL09} proposed a multi-field correlated topic model by relaxing the assumption of using common set of topics globally among all documents, which can also be applied to the probit model to enrich the comprehensiveness of structural relationships between topics . Another potential direction would be to enhance the scalability of the model. Currently we used a simple distributed algorithm proposed by \cite{yao2009efficient} and \cite{Newman:2009:DAT:1577069.1755845} for efficient Gibbs sampling. The architecture for topic models presented by \cite{smola2010architecture} can be further utilized to reduce the computational complexity substantially while delivering comparable performance. Furthermore, a novel sampling method involving the Gibbs Max-Margin Topic \cite{journals/corr/ZhuCPZ13} will further improve the computational efficiency.\\





\bibliographystyle{chicago}
\bibliography{Probit_Normal_Correlated_Topic_Model}

\begin{thebibliography}{}

\bibitem[\protect\citeauthoryear{Albert and Chib}{Albert and
  Chib}{1993}]{AlbertChib93}
Albert, J.~H. and S.~Chib (1993).
\newblock Bayesian analysis of binary and polychotomous response data.
\newblock {\em Journal of the American Statistical Association\/}~{\em
  88\/}(422), 669--679.

\bibitem[\protect\citeauthoryear{Blei and Lafferty}{Blei and
  Lafferty}{2006}]{Blei06correlatedtopic}
Blei, D.~M. and J.~D. Lafferty (2006).
\newblock Correlated topic models.
\newblock In {\em In Proceedings of the 23rd International Conference on
  Machine Learning}, pp.\  113--120. MIT Press.

\bibitem[\protect\citeauthoryear{Blei, Ng, Jordan, and Lafferty}{Blei
  et~al.}{2003}]{Blei03latentdirichlet}
Blei, D.~M., A.~Y. Ng, M.~I. Jordan, and J.~Lafferty (2003).
\newblock Latent dirichlet allocation.
\newblock {\em Journal of Machine Learning Research\/}~{\em 3}, 2003.

\bibitem[\protect\citeauthoryear{Chen, Zhu, Wang, Zheng, and Zhang}{Chen
  et~al.}{2013}]{Tsinghua:2013:1}
Chen, J., J.~Zhu, Z.~Wang, X.~Zheng, and B.~Zhang (2013).
\newblock Scalable inference for logistic-normal topic models.
\newblock In C.~Burges, L.~Bottou, M.~Welling, Z.~Ghahramani, and K.~Weinberger
  (Eds.), {\em Advances in Neural Information Processing Systems 26}, pp.\
  2445--2453. Curran Associates, Inc.

\bibitem[\protect\citeauthoryear{Grün and Hornik}{Grün and
  Hornik}{}]{Grün:Hornik:2011:JSSOBK:v40i13}
Grün, B. and K.~Hornik.
\newblock topicmodels: An r package for fitting topic models.
\newblock {\em Journal of Statistical Software\/}~{\em 40\/}(13), 1--30.

\bibitem[\protect\citeauthoryear{Johndrow, Lum, and Dunson}{Johndrow
  et~al.}{2013}]{Johndrow:2013:1}
Johndrow, J., K.~Lum, and D.~B. Dunson (2013).
\newblock Diagonal orthant multinomial probit models.
\newblock In {\em JMLR Proceedings}, Volume~31 of {\em AISTATS}, pp.\  29--38.
  JMLR.

\bibitem[\protect\citeauthoryear{Mimno, Wallach, and Mccallum}{Mimno
  et~al.}{2008}]{Mimno_gibbssampling}
Mimno, D., H.~M. Wallach, and A.~Mccallum (2008).
\newblock Gibbs sampling for logistic normal topic models with graph-based
  priors.

\bibitem[\protect\citeauthoryear{Newman, Asuncion, Smyth, and Welling}{Newman
  et~al.}{2009}]{Newman:2009:DAT:1577069.1755845}
Newman, D., A.~Asuncion, P.~Smyth, and M.~Welling (2009, December).
\newblock Distributed algorithms for topic models.
\newblock {\em J. Mach. Learn. Res.\/}~{\em 10}, 1801--1828.

\bibitem[\protect\citeauthoryear{Salomatin, Yang, and Lad}{Salomatin
  et~al.}{2009}]{DBLP:conf:sdm:SalomatinYL09}
Salomatin, K., Y.~Yang, and A.~Lad (2009).
\newblock Multi-field correlated topic modeling.
\newblock In {\em Proceedings of the {SIAM} International Conference on Data
  Mining, {SDM} 2009, April 30 - May 2, 2009, Sparks, Nevada, {USA}}, pp.\
  628--637.

\bibitem[\protect\citeauthoryear{Smola and Narayanamurthy}{Smola and
  Narayanamurthy}{2010}]{smola2010architecture}
Smola, A. and S.~Narayanamurthy (2010).
\newblock An architecture for parallel topic models.
\newblock In {\em VLDB}.

\bibitem[\protect\citeauthoryear{Yao, Mimno, and McCallum}{Yao
  et~al.}{2009}]{yao2009efficient}
Yao, L., D.~Mimno, and A.~McCallum (2009).
\newblock Efficient methods for topic model inference on streaming document
  collections.
\newblock In {\em KDD}.

\bibitem[\protect\citeauthoryear{Zhu, Chen, Perkins, and Zhang}{Zhu
  et~al.}{2013}]{journals/corr/ZhuCPZ13}
Zhu, J., N.~Chen, H.~Perkins, and B.~Zhang (2013).
\newblock Gibbs max-margin topic models with data augmentation.
\newblock {\em CoRR\/}~{\em abs/1310.2816}.

\end{thebibliography}
\end{document}